\documentclass[twoside]{article}

\usepackage[accepted]{aistats2021}
\usepackage{microtype}
\usepackage{hyperref}
\usepackage{url}

\usepackage{color}
\usepackage{amsmath,amsfonts,amsthm,bm}
\usepackage{amssymb}
\usepackage{pifont}

\usepackage{multirow}
\usepackage{multicol}
\usepackage{graphicx}

\usepackage{wrapfig}
\usepackage{lipsum}  

\usepackage{algorithm}
\usepackage{algorithmic}

\usepackage{caption} 
\usepackage{subcaption}
\usepackage{graphicx}
\usepackage{wrapfig}

\usepackage{epstopdf}
\usepackage{enumerate}

\usepackage{float}
\usepackage{tabularx}
\usepackage{makecell}
\usepackage{enumitem}
\usepackage{arydshln}
\usepackage{footmisc}

\usepackage{booktabs,tabularx}

\newtheorem{definition}{Definition}

\newcommand{\xv}{{\mathbf x}}

\newcommand{\Xv}{{\mathbf X}}

\newcommand{\Xc}{{\mathcal X}}
\newcommand{\Yc}{{\mathcal Y}}
\newcommand{\Zc}{{\mathcal Z}}
\newcommand{\Vc}{{\mathcal V}}

\newcommand{\loss}{{\mathcal L}}
\newcommand{\data}{{\mathcal D}}

\newcommand{\E}{{\mathbb{E}}}

\newcommand{\Norm}{{\mathcal{N}}}

\newcommand{\proposed}{{DeepIMV}}

\newcolumntype{L}[1]{>{\raggedright\let\newline\\\arraybackslash\hspace{0pt}}m{#1}}
\newcolumntype{C}[1]{>{\centering\let\newline\\\arraybackslash\hspace{0pt}}m{#1}}
\newcolumntype{R}[1]{>{\raggedleft\let\newline\\\arraybackslash\hspace{0pt}}m{#1}}

\newcommand{\defeq}{\overset{\mathrm{def}}{=}}

\newcommand{\cmark}{\ding{51}}%
\newcommand{\xmark}{\ding{55}}%

\newsavebox\CBox 
\def\textBF#1{\sbox\CBox{#1}\resizebox{\wd\CBox}{\ht\CBox}{\textbf{#1}}}

\begin{document}

%
\runningtitle{A Variational IB Appproach to Multi-Omics Data Integration}

%

\twocolumn[

\aistatstitle{A Variational Information Bottleneck Approach to\\ Multi-Omics Data Integration}

\aistatsauthor{ Changhee Lee \And Mihaela van der Schaar}

\aistatsaddress{ University of California, Los Angeles \And  University of Cambridge \\ University of California, Los Angeles \\ Alan Turing Institute} ]

\begin{abstract}
    Integration of data from multiple omics techniques is becoming increasingly important in biomedical research. Due to non-uniformity and technical limitations in omics platforms, such integrative analyses on multiple omics, which we refer to as views, involve learning from incomplete observations with various view-missing patterns. This is challenging because i) complex interactions within and across observed views need to be properly addressed for optimal predictive power and ii) observations with various view-missing patterns need to be flexibly integrated. 
	To address such challenges, we propose a deep variational information bottleneck (IB) approach for \textit{incomplete multi-view} observations. 
	Our method applies the IB framework on marginal and joint representations of the observed views to focus on intra-view and inter-view interactions that are relevant for the target. Most importantly, by modeling the joint representations as a product of marginal representations, we can efficiently learn from observed views with various view-missing patterns.
	Experiments on real-world datasets show that our method consistently achieves gain from data integration and outperforms state-of-the-art benchmarks.
\end{abstract}

\section{Introduction}
Technological advances in high-throughput biology enable integrative analyses that use information across multiple omics layers -- including genomics, epigenomics, transcriptomics, proteomics, and metabolomics -- to deliver more comprehensive understanding in biological systems \cite{MultiOmics_ref3,MultiOmics_ref4}. 
Unfortunately, due to limitations of experimental designs or compositions from different data platforms (e.g., TCGA\footnote{\url{http://cancergenome.nih.gov/}}), integrated samples commonly have one or more entirely missing omics with various missing patterns.
Learning from such incomplete observations is challenging. Discarding samples with missing omics greatly reduces sample sizes (especially when integrating many omics layers) \cite{MultiOmics_ref1} and simple mean imputation can seriously distort the marginal and joint distribution of the data \cite{MultiOmics_ref2}.

In this paper, we model multi-omics data integration as learning from \textit{incomplete multi-view observations} where we refer to observations from each omics data as \textit{views} (e.g., DNA copy number and mRNA expressions).
However, direct application of the existing multi-view learning methods does not address the key challenges of handling missing views in integrating multi-omics data.
This is because these methods are typically designed for complete-view observations assuming that all the views are available for every sample \cite{Li:19}. 
Therefore, we set our goal to develop a model that not only learns complex intra-view and inter-view interactions that are relevant for target tasks but also flexibly integrates observed views regardless of their view-missing patterns in a single framework.

\textbf{Contribution.} 
To this goal, we propose a deep variational information bottleneck (IB) approach for incomplete multi-view observations, which we refer to as \proposed. 
Our method consists of four network components: a set of \textit{view-specific encoders}, a set of \textit{view-specific predictors}, a \textit{product-of-experts} (PoE) module, and a \textit{multi-view predictor}.
More specifically, for flexible integration of the observed views regardless of the view-missing patterns, we model the \textit{joint} representations as a PoE over the marginal representations, which are further utilized by the multi-view predictor.
Thus, the joint representations combine both common and complementary information across the observed views. 
The entire network is trained under the IB principle \cite{Tishby:99,Alemi:17} which encourages the marginal and joint representations to focus on intra-view and inter-view interactions that are relevant to the target, respectively. 
Experiments on real-world datasets show that our method consistently achieves gain from data integration and significantly outperforms state-of-the-art benchmarks with respect to measures of prediction performance.


\section{Related Works}
\subsection{Multi-View Learning}

\textbf{Complete Multi-View Observations.} 
To utilize information across multiple views, a variety of methods have been proposed in recent years.
Canonical correlation analysis (CCA) \cite{Hotelling:36} and its kernel or deep learning extensions \cite{Akaho:06,Andrew:13,Wang:15} are representative methods which aim to learn a common (latent) space that is shared between the two views such that their canonical correlation is maximized in a purely unsupervised fashion.
The learned representations can then be used for supervised learning as a downstream task. Meanwhile, some existing methods \cite{Diethe:08,Kan:12,Jia:19} have focused on fully utilizing the label information when seeking for the common (latent) space that are not only shared between the views but also discriminative for the target.  
Although these methods have shown promising performance in different applications \cite{Li:19}, they are designed for complete-view observations. 


\textbf{Incomplete Multi-View Observations. } 
Most of the existing methods require multi-stage training -- that is, constructing complete multi-view observations based on imputation methods \cite{Cai:16,Tran:17} and then training a multi-view model -- or relying on auxiliary inference steps to generate the missing views \cite{Amini:09,Tsai:19,Shi:19}. 
There are a few methods that flexibly handle incomplete multi-view observations. 
Authors in \cite{Wu:18} introduced a generative model that utilizes latent factorization enabling cross-view generation without requiring multi-stage training regimes or additional inference steps.
Matrix factorization was extended in \cite{MOFA} to learn low-dimensional representations that capture the joint aspects across different views. 
However, these methods are trained in a purely unsupervised fashion. Thus, while information relevant to the reconstruction of views will be well-captured in learned representations, information relevant to the target task may be lost.

Our work is most closely related to CPM-Nets \cite{Zhang:19}.
Both methods aim to find representations in a common space for supervised learning of predicting the target. A notable distinction from CPM-Nets is how we integrate incomplete multi-view observations.
In CPM-Nets, the authors directly learn the mapping from latent representations to the original views without utilizing any encoder structure.\footnote{The ``encoding networks'' in \cite{Zhang:19} denote networks that reconstruct original views from latent representations; there are no such networks that map original views to latent representations (in our context, encoders).}
Instead, for all the training samples, the corresponding latent representations are randomly initialized and, then, iteratively updated to minimize the reconstruction and classification losses.
We conjecture that this approach has two limitations:
First, the method relies on reconstruction to find latent representations of the testing samples, which makes it difficult to capture task-relevant information.
Second, random initialization means that there are no inherent relations across the latent representations of the training samples. Thus, all training samples must be updated at the same time, which increases the memory burden during training.
In contrast, in \proposed, we utilize posterior factorization for flexible integration of observed views and focus latent representations to be task-relevant, thereby effectively capturing the complementary information for predicting the target.
In addition, CPM-Nets is designed only for classification tasks; thus, extension to regression tasks can lose the advantages of cluster-friendly representations.

\subsection{Information Bottleneck}
The information bottleneck (IB) principle \cite{Tishby:99,Amini:09,Achille:18} is an information-theoretic approach that defines the intuitive idea of what a \textit{task-relevant representation} is in terms of the fundamental trade-off between having a concise representation and one that provides good prediction power.
The IB framework \cite{Alemi:17} has been recently studied to address multi-view problems \cite{Wang:19,Federici:20}. Authors in \cite{Wang:19} combined marginal representations from each view using an auxiliary layer into joint representation on which the IB principle is applied. However, this work cannot handle missing views both during training and testing. 
In \cite{Federici:20}, the IB principle was extended to the two-view unsupervised setting to learn robust representations that are common to both views. Thus, this is not applicable to multi-omics data integration where the goal is to flexibly integrate incomplete views (i.e., observations from different omics layers), that often contain both common and complementary information, in a supervised fashion. 



%
%

\begin{figure*}[t!]
	\centering 
	\includegraphics[width=0.80\linewidth, trim={0cm 0cm 0cm 0cm}, clip]{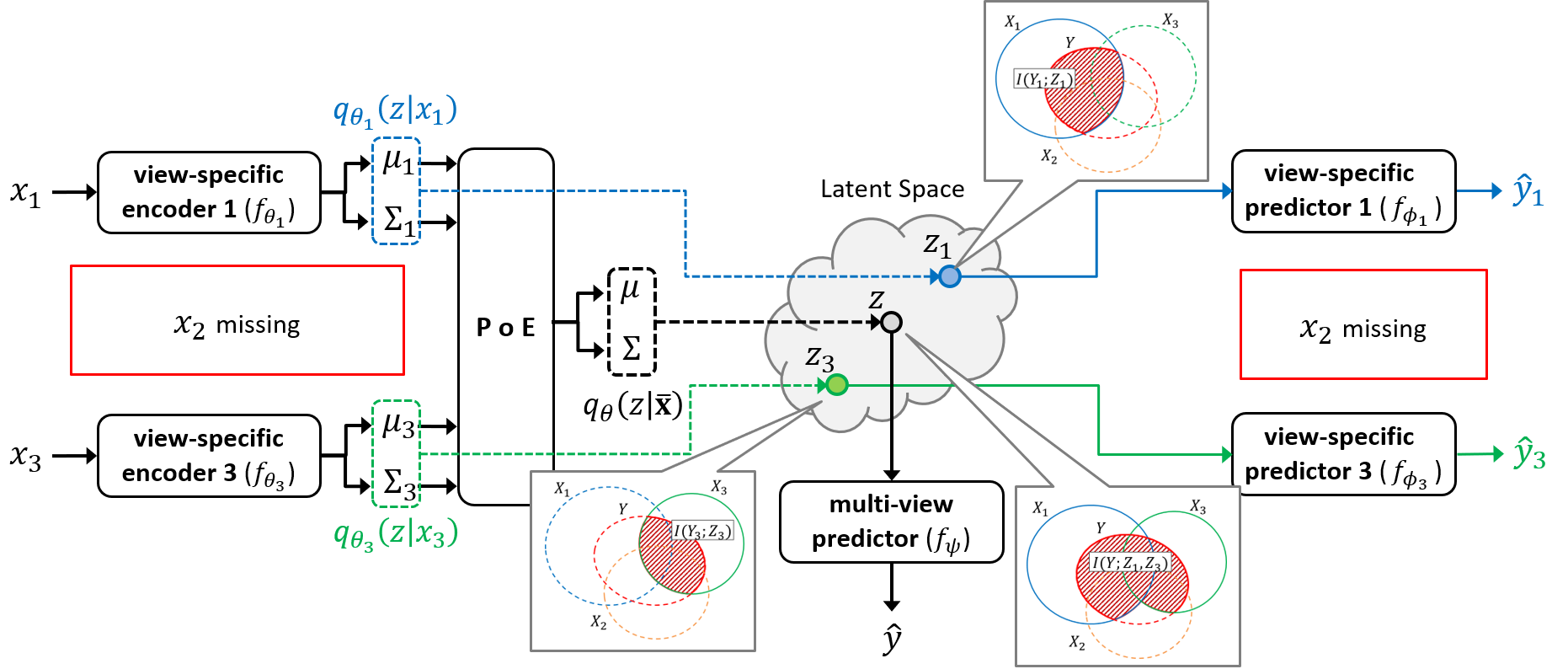} \vspace{-1mm}
	\caption{An illustration of the proposed network architecture with $V=3$ views. For an illustration purpose, we assume that the current sample has the second view missing, i.e., $x_{2}=\varnothing$. Here, dotted lines correspond to drawing samples in the latent representation from the respective distributions, and green and blue colored lines indicate marginal representations and the corresponding predictions.}
	\label{fig:network_architecture} \vspace{-2mm}
\end{figure*}
\section{Incomplete Multi-View Problem}

The presence of missing views remains an inevitable and prevalent problem in multi-omics data integration. To address this, we start off by defining such an integrative analysis as an \textit{incomplete multi-view problem} where some of the views may be missing with arbitrary view-missing patterns.

\textbf{Notation.}
Let $X_{v} \in \Xc_{v}$ be a random variable for the $d_{v}$-dimensional input feature from the $v$-th view and $Y \in \Yc$ be a random variable for the output label. 
We say the $v$-th view is missing if $x_{v} = \varnothing$ and is available if $x_{v} \in \Xc_{v}$ where $x_{v}$ is a realization of $X_{v}$. 
To incorporate with arbitrary missing views, we denote a set of observed views by $\Vc = \{ v: x_{v} \neq \varnothing \} \subseteq [V]$ which we refer to as a \textit{view-missing pattern}. Here, $[V]$ is the complete set of available views. Then, we can finally define a random variable for a \textit{multi-view observation} as $\bar{\Xv} = (X_{v}: v\in\Vc)$; we say the observation is \textit{incomplete} when $\Vc\subset [V]$ and \textit{complete} when $\Vc=[V]$. Throughout the paper, we will often use lower-case letters to denote realizations of a random variable.

\begin{definition}
	\textnormal{\textbf{(Incomplete Multi-View Problem)}} 
	Suppose that we have a training dataset $\data = \{\bar{\xv}^{n}, y^{n}\}_{n=1}^{N}$ which contains one or more sample(s) with missing views, i.e., $\Vc^{n} \subset [V]$ for some $n$.
	Then, we define an incomplete multi-view problem as a supervised learning problem of predicting target $y$ -- that is, $C$-classification for $\Yc = [C]$ and regression for $\Yc = \mathbb{R}$ -- for a new multi-view observation $\bar{\xv}$ with an arbitrary view-missing pattern.\footnote{Note that a similar definition has been introduced in \cite{Zhang:19} for classification tasks.}	
\end{definition}

Solving an incomplete multi-view problem involves two main challenges: 
First, we want to learn representations in a \textit{common space} that leverages both \textit{marginal} and \textit{joint} aspects of the observed views for prediction. 
Second, the learned representations must flexibly integrate incomplete observations with various view-missing patterns in an unified framework.




\section{Method: \proposed}
To address such challenges, we propose a deep variational information bottleneck approach, which we refer to as \proposed\footnote{Source code is available at \url{https://github.com/chl8856/DeepIMV}.}, that consists of four network components as illustrated in Figure \ref{fig:network_architecture}: 
\vspace{-2mm}
\begin{itemize}[leftmargin=1.5em]
    \item a set of $V$ \textit{view-specific encoders}, parameterized by $\theta=(\theta_{1}, \cdots, \theta_{V})$, each of which stochastically maps observations from each individual view into a common latent space;\vspace{-1.mm} 
    \item a \textit{product-of-experts} (PoE) module that integrates the marginal latent representations into a joint latent representation in the common space;  \vspace{-1.mm}
	\item a \textit{multi-view predictor}, parameterized by $\psi$, which provides label predictions based on the joint representations; and \vspace{-1.mm}
    \item a set of $V$ \textit{view-specific predictors}, parameterized by $\phi=(\phi_{1}, \cdots, \phi_{V})$, each of which provides label predictions based on the marginal latent representations determined by the corresponding view-specific encoders.\vspace{-1.mm}
\end{itemize} \vspace{-1.mm}
Throughout the remainder of the paper, we will often use $f$ to denote a deterministic mapping and use $q$ to denote a stochastic mapping.



\subsection{Toward Task-Relevant Representations} \label{sec:relevance}
Finding \textit{task-relevant representations} in a common latent space that contains both \textit{marginal} and \textit{joint} aspects of the observations is crucial for solving the incomplete multi-view problem.
To this goal, we apply the IB principle \cite{Alemi:17,Achille:18} since it aims at learning task-relevant representations by discarding as much information about the input as possible that is irrelevant to the target task and thereby encouraging the predictor to be robust to overfitting.


Define $\Zc$ be the common latent space. 
We consider the \textit{marginal representation} $Z_{v} \in \Zc$ be a stochastic encoding of $X_{v}$, i.e., $q_{\theta_{v}}\!(Z_{v}|X_{v})$, which is defined by the $v$-th view-specific encoder for $v\in\Vc$. 
Similarly, we consider the \textit{joint representation} $Z \in \Zc$ be a stochastic encoding of $\bar{\Xv}$, which is defined by the encoder block $q_{\theta}(Z|\bar{\Xv})$ that combines outputs of the $V$ view-specific encoders.
Then, given a latent representation $z\in\Zc$ drawn from $q_{\theta}(z|\bar{\xv})$, the multi-view predictor estimates the target label $Y$ based on the distribution defined as $q_{\psi}(Y|z)$. 
To learn \textit{joint} aspects of the observed views for predicting the target $Y$, we apply the IB principle on $Z$ based on the following loss:
\begin{equation}\label{eq:loss_IB_joint}
\begin{split}
	\loss_{\text{IB-J}}^{\theta,\psi}(\bar{\Xv}, Y) &= - I(Y; Z) + \beta I(\bar{\Xv}; Z) \\
	&\approx \E_{\bar{\xv},y\sim p(\bar{\xv},y)} \E_{z\sim q_{\theta}(z|\bar{\xv})}\! \Big[\!-\log q_{\psi}(y|z) \Big] \\
	&+ \beta \E_{\bar{\xv}\sim p(\bar{\xv})}\! \Big[ KL\big(q_{\theta}(Z|\bar{\xv}) \big\| q(Z)\big) \Big]
\end{split}
\end{equation}
where $\beta \geq 0$ is a coefficient chosen to balance between the two information quantities. Here, $KL\big(p(Z) \big\| r(Z)\big)$ denotes the Kullback-Leibler (KL) divergence between the two distributions $p(Z)$ and $r(Z)$. 
Detailed derivations can be found in the Supplementary Material.


\subsection{Product-of-Experts for Incomplete Views} \label{sec:factorization}
One question is outstanding. How should we design the joint representation such that it integrates the marginal representations of the observed views with arbitrary view-missing patterns?
Our solution is to use a product-of-experts (PoE) that factorizes the joint posterior $q_{\theta}(Z|\bar{\Xv})$ into a \textit{product} of the marginal posteriors $q_{\theta_{v}}\!(Z|X_{v})$ for $v\in\Vc$. 
Formally, the joint posterior can be defined as the following \cite{Wu:18}:
\begin{equation} \label{eq:product_of_posteriors}
    p(z|\bar{\xv}) \approx C \cdot p(z) \prod_{v\in\Vc} q_{\theta_{v}}(z|x_{v}) \defeq q_{\theta}(z|\bar{\xv})
\end{equation}
where $C$ is a normalizing constant.

Factorizing the joint posterior using a PoE is promising for solving an incomplete multi-view problem compared to that using a mixture-of-experts (MoE), i.e. $q_{\theta}(z|x_{1},\cdots, x_{V}) = \sum_{v=1}^{V} \alpha_{v} \cdot q_{\theta_{v}}(z|x_{v}) $: First, by employing PoE, we can simply ignore missing views when finding the joint representation $Z$ of an input $\bar{\Xv}$ regardless of its view-missing patterns. Hence, we can fully utilize samples with incomplete views without discarding them (or applying view-completion in advance) for training \cite{Shi:19}, and can avoid auxiliary inference steps of generating the missing views \cite{Amini:09,Tsai:19} during both training and testing. 
Second, aside from its flexibility, a PoE (i.e., the joint posterior) can produce a much sharper distribution than the individual experts (i.e., the marginal posteriors) allowing for them to be specialized in a particular aspect of the target task \cite{Hinton:02}. 
This is a desirable property for multi-omics data integration where individual views often contain view-specific (complementary) information or uneven amount of information about the target.

\textbf{Computing the Joint Posterior.}
Suppose that we utilize marginal posteriors of the form $q_{\theta_{v}}\!(z|x_{v}) = \Norm(z|\mu_{v}, \Sigma_{v})$ where $(\mu_{v}, \Sigma_{v}) = f_{\theta_{v}}(x_{v})$ and use the prior of the form $p(z) = \Norm(z|\mu_{0}, \Sigma_{0})$ (typically, a spherical Gaussian). Because a product of Gaussian is itself Gaussian \cite{Cao:14}, we can derive the joint posterior as $q_{\theta}(z|\bar{\xv}) = \Norm(z|\mu, \Sigma)$ where 
$\mu = (\mu_{0}\Sigma_{0}^{-1} + \sum_{v\in\Vc} \mu_{v} \Sigma_{v}^{-1} )(\Sigma_{0}^{-1} + \sum_{v\in\Vc} \Sigma_{v}^{-1} )^{\!-1}$ and $\Sigma = (\Sigma_{0}^{-1} + \sum_{v\in\Vc} \Sigma_{v}^{-1} )^{\!-1}$. 
Hence, we can efficiently compute the joint posterior of the incomplete multi-view observations in terms of the available marginal posteriors.

\subsection{Building View-Specific Expertise}
However, training a PoE appears to be difficult, requiring artificial sub-sampling of the observed views \cite{Wu:18} or applying variants of contrastive divergence \cite{Hinton:02} in order to ensure that
the individual views are learnt faithfully. 
To mitigate such an issue, we instead introduce a set of $V$ view-specific predictors and apply IB principle on the marginal representations in order to allow each encoder to build view-specific expertise for predicting the target.

Formally, for each view, given the latent representation $z_{v}\in\Zc$ drawn from $q_{\theta_{v}}\!(z_{v}|x_{v})$, the view-specific predictor estimates $Y_{v}$, which is a random variable for the target that can be solely described by the corresponding view $X_{v}$, based on the distribution defined as $q_{\phi_{v}}(Y_{v}|z_{v})$. 
Then, we apply the IB principle on the marginal representations to capture view-specific aspects of the observed views for predicting the target based on the following loss:
\begin{equation} \label{eq:loss_IB_marginal}
\begin{split}
&\loss_{\text{IB-M}}^{\theta_{v},\phi_{v}}(X_{v}, Y_{v}) = -I(Y_{v}; Z_{v}) + \beta_{v} I(X_{v}; Z_{v}) \\
&~~~~~~~~~\approx \E_{x_{v},y\sim p(x_{v},y)} \E_{z\sim q_{\theta_{v}}\!(z|x_{v})}\! \Big[\!-\!\log q_{\phi_{v}}\!(y|z) \Big] \\
&~~~~~~~~~+\beta_{v} \E_{x_{v}\sim p(x_{v})}\! \Big[ KL\big(q_{\theta_{v}}\!(Z_{v}|x_{v}) \big\| q(Z_{v})\big) \Big]  
\end{split}
\end{equation}
where $\beta_{v} \geq 0$ is a balancing coefficient. 
Minimizing \eqref{eq:loss_IB_marginal} encourages $Z_{v}$ to become a minimal suffcient statistics of $X_{v}$ for $Y_{v}$ \cite{Tishby:10}, enforcing the marginal representations to learn the view-specific (possibly complementary) aspects of the target, which eventually ease the training of PoE.


\subsection{Training}
We train the overall network -- the view-specific encoder-predictor pairs $(\theta, \phi)$ and the final predictor $(\psi)$ -- by minimizing a combination of the marginal and joint IB losses: 
\begin{equation}
\loss_{\text{Total}}^{\theta,\phi,\psi}(\bar{\Xv}, Y) = \loss_{\text{IB-J}}^{\theta,\psi}(\bar{\Xv}, Y) +  \alpha  \sum_{v\in\Vc} \loss_{\text{IB-M}}^{\theta_{v},\phi_{v}}(X_{v},Y_{v})
\end{equation}
where $\alpha \geq 0$ is a hyper-parameter that trades off between the two losses. The pseudo-code of \proposed~is provided in Algorithm \ref{alg:pseudo_code}.\footnote{Here, we assume a $C$-classification task and slightly abuse notation; we will write $y$ to denote a one-hot vector and write $y[c]$ to denote the $c$-th element of $y$.}


\begin{algorithm}[t!]
	\caption{Pseudo-code for training \proposed}
	\label{alg:pseudo_code}
	\begin{algorithmic}
		\footnotesize
		\STATE {\bfseries Input:} $\data = \{\bar{\xv}^{n}, y^{n}\}_{n=1}^{N}$, learning rate $\eta$, mini-batch size $n_{mb}$, coefficients $\alpha, \beta, \{\beta_{v}\}_{v=1}^{V}$, and $\mu_{0}$, $\Sigma_{0}$
		\STATE {\bfseries Output:} \proposed~parameters $(\theta, \phi, \psi)$
		\STATE {Initialize parameters $(\theta, \phi, \psi)$ via Xavier Initializer}
	
		\REPEAT
		\STATE Sample a mini-batch: $\{\bar{\xv}^{n}, y^{n}\}_{n=1}^{n_{mb}} \sim \data$
		\FOR{$n=1,\cdots, n_{mb}$}
		\FOR{$v \in \Vc^{n}$}
		\STATE $\mu_{v}^{n}, \Sigma_{v}^{n} \leftarrow f_{\theta_{v}}(x_{v}^{n})$
		\STATE $z_{v}^{n} \leftarrow \mu_{v}^{n} + \epsilon\cdot\Sigma_{v}^{n}$~~where $\epsilon\sim \Norm(0,1)$
		\STATE$\hat{y}_{v}^{n} \leftarrow f_{\phi_{v}}(z_{v}^{n})$
		\ENDFOR
		\STATE $\mu^{n}, \Sigma^{n} \leftarrow PoE\big((\mu_{v}^{n}, \Sigma_{v}^{n})_{v\in \Vc^{n}}\big)$
		\STATE $z^{n} \leftarrow \mu^{n} + \epsilon\cdot\Sigma^{n}$~~where $\epsilon\sim \Norm(0,1)$
		\STATE $\hat{y}^{n} \leftarrow f_{\psi}(z^{n})$
		\ENDFOR
		\STATE $\hat{\loss}_{\text{IB-J}}^{\theta, \psi} = \frac{1}{n_{mb}} \sum_{n=1}^{n_{mb}} \big(-\sum_{c=1}^{C}y^{n}[c]\log(\hat{y}^{n}[c])$
		\STATE ~~~~~~~~~~~$+~~\beta \cdot KL(\Norm(\mu^{n},\Sigma^{n})||\Norm(\mu_{0},\Sigma_{0})) \big)$
		\STATE $\hat{\loss}_{\text{IB-M}}^{\theta,\phi} = \frac{1}{n_{mb}} \sum_{n=1}^{n_{mb}} \sum_{v\in\Vc^{n}} \big(-\sum_{c=1}^{C}y^{n}[c]\log(\hat{y}_{v}^{n}[c])$
		\STATE ~~~~~~~~~~~$+~~\beta_{v} \cdot KL(\Norm(\mu_{v}^{n},\Sigma_{v}^{n})||\Norm(\mu_{0},\Sigma_{0})) \big)$
		\STATE $(\theta, \phi, \psi) \leftarrow (\theta, \phi, \psi) - \eta \nabla_{(\theta, \phi, \psi)} \big(\hat{\loss}_{\text{IB-J}}^{\theta, \psi} + \alpha \hat{\loss}_{\text{IB-M}}^{\theta, \phi}\big)$
		\UNTIL convergence
	\end{algorithmic}
\end{algorithm}

\section{Experiments}
Throughout the experiments, we evaluate different multi-view learning methods on two real-world multi-omics datasets that were collected by the Cancer Genome Atlas (TCGA)\footnote{\url{ https://www.cancer.gov/tcga}} and by the Cancer Cell Line Encyclopedia (CCLE) \cite{CCLE_ref} in the context of integrating multi-omics observations for predicting 1-year mortality and drug sensitivity of cancer cells, respectively.

\textbf{Benchmarks. } 
We compare \proposed~with 2 baselines and 6 state-of-the-art multi-view learning methods: The baselines include a pre-integration method that simply concatenates observations from multiple views (denoted as \textbf{Base1}) and a post-integration method that integrates predictions of individual predictors trained on each view as an ensemble (denoted as \textbf{Base2}). 
The multi-view learning methods that assume complete multi-view observations include \textbf{GCCA} \cite{Kettenring:71}, \textbf{DCCA} \cite{Andrew:13}, and \textbf{DCCAE} \cite{Wang:15}, and the methods that flexibly integrate incomplete multi-view observations include \textbf{MVAE} \cite{Wu:18}, \textbf{CPM-Nets} \cite{Zhang:19}, and \textbf{MOFA} \cite{MOFA}. Table \ref{table:benchmarks} summarizes the key characteristics. 
It is worth highlighting that i) for the methods that cannot handle incomplete multi-view observations, we use mean values (and further utilize the reconstructed inputs of MVAE in Tables \ref{table:imputed_results_tcga} and \ref{table:imputed_results_ccle}) to impute missing views of training and testing samples, and ii) for the methods that do not utilize label information, we train a multi-layer perceptron (MLP) based on the learned representations as a downstream task. 
\begin{table}[h!]
	\centering 
	\caption{Comparison table.} \vspace{-2mm}
	\label{table:benchmarks}
	\small
	\begin{tabular}{c c c}
	\toprule 
		\textbf{Methods} &\textbf{\makecell{Task\\Oriented}} & \textbf{\makecell{Incomplete\\Views}} \\ \midrule
		Base1    & {\color{blue}\cmark}  & \xmark \\
		Base2    & {\color{blue}\cmark}  & \xmark \\
		GCCA     & \xmark  & \xmark \\
		DCCA     & \xmark  & \xmark \\
		DCCAE    & \xmark  & \xmark \\
		MVAE     & \xmark  & {\color{blue}\cmark} \\
		CPM-Nets & {\color{blue}\cmark}  & {\color{blue}\cmark} \\
		MOFA     & \xmark  & {\color{blue}\cmark} \\
		\proposed & {\color{blue}\cmark}  & {\color{blue}\cmark} \\
		\bottomrule
	\end{tabular} 
\end{table}


To focus our experiments on the integrative analysis and to overcome ``curse-of-dimensionality'' in the high-dimensional multi-omics data, we extracted low-dimensional representations (i.e., 100 features) using the kernel-PCA (with polynomial kernels) on each view \cite{Shiokawa:18} to train all the multi-view learning methods except for MOFA, which is well-known for
capturing sparse factors across multiple views. Please see the Supplementary Material for more details.

Implementations details and sensitivity analyses on the hyper-parameters $(\alpha, \beta)$ of \proposed~and details on the benchmarks can be found in the Supplementary Material. Throughout the experiments, all the results are reported based on 10 and 100 random 64/16/20 train/validation/test splits for the TCGA dataset and the CCLE dataset, respectively.

\begin{table*}[t!]
	\scriptsize
	\centering
	\caption{Comparison of the AUROC performance (mean $\pm$ 95\%-CI) for (\textit{complete}) the multi-view learning methods that are trained with only complete multi-view samples and (\textit{incomplete}) those that are trained with both complete and incomplete multi-view samples. The values are reported by varying the number of observed views from $|\Vc^{n}| = 1$ to $|\Vc^{n}| = 4$ during testing.}
	\label{table:tcga_results}
	\begin{tabular}{| c | c c | c c | c c | c c |}
		\hline 
		\multirow{2}{*}{\textbf{Methods}} &\multicolumn{2}{c|}{\textbf{\underline{~~~~~\textit{1 View}~~~~~}}} & \multicolumn{2}{c|}{\textbf{\underline{~~~~~\textit{2 Views}~~~~~}}} & \multicolumn{2}{c|}{\textbf{\underline{~~~~~\textit{3 Views}~~~~~}}}  &
		\multicolumn{2}{c|}{\textbf{\underline{~~~~~\textit{4 Views}~~~~~}}} \\ 
		&\textit{complete} &\textit{incomplete} &\textit{complete} &\textit{incomplete} &\textit{complete} &\textit{incomplete} &\textit{complete} &\textit{incomplete}\\ \hline 
\textbf{Base1}	&0.660$\pm$0.04&0.675$\pm$0.02&0.722$\pm$0.03&0.739$\pm$0.02&0.750$\pm$0.02&0.765$\pm$0.02&0.766$\pm$0.02&0.781$\pm$0.01 \\
\textbf{Base2}	&\color{blue}\textBF{0.711$\pm$0.02}&0.717$\pm$0.02&0.746$\pm$0.01&0.766$\pm$0.00&0.767$\pm$0.02&0.775$\pm$0.01&\color{blue}\textBF{0.783$\pm$0.02}&0.790$\pm$0.01 \\
\textbf{GCCA}	&0.680$\pm$0.02&0.650$\pm$0.03&0.737$\pm$0.02&0.737$\pm$0.03&0.764$\pm$0.01&0.769$\pm$0.02&\color{blue}\textBF{0.783$\pm$0.01}&0.792$\pm$0.01 \\
\textbf{DCCA}	&0.702$\pm$0.01&0.638$\pm$0.03&0.745$\pm$0.03&0.761$\pm$0.02&0.758$\pm$0.02&0.775$\pm$0.01&0.776$\pm$0.02&0.784$\pm$0.01 \\
\textbf{DCCAE}	&0.623$\pm$0.04&0.605$\pm$0.04&0.747$\pm$0.03&0.763$\pm$0.01&0.774$\pm$0.02&0.775$\pm$0.01&0.776$\pm$0.02&0.778$\pm$0.02 \\
\textbf{MVAE}	&0.592$\pm$0.05&0.589$\pm$0.04&0.677$\pm$0.02&0.674$\pm$0.02&0.731$\pm$0.02&0.730$\pm$0.01&0.774$\pm$0.01&0.781$\pm$0.01 \\
\textbf{CPM-Nets}	&0.700$\pm$0.02&0.709$\pm$0.01&0.748$\pm$0.02&0.761$\pm$0.02&0.766$\pm$0.01&0.771$\pm$0.01&0.781$\pm$0.01&0.788$\pm$0.01 \\
\textbf{MOFA}	&0.681$\pm$0.03&0.646$\pm$0.01&0.732$\pm$0.01&0.734$\pm$0.01&0.756$\pm$0.01&0.764$\pm$0.02&0.781$\pm$0.02&0.785$\pm$0.02  \\
\textbf{DeepIMV}	&0.701$\pm$0.02&\color{blue}\textBF{0.724$\pm$0.02}&\color{blue}\textBF{0.757$\pm$0.02}&\color{blue}\textBF{0.772$\pm$0.01}&\color{blue}\textBF{0.776$\pm$0.01}&\color{blue}\textBF{0.791$\pm$0.01}&\color{blue}\textBF{0.783$\pm$0.01}&\color{blue}\textBF{0.801$\pm$0.01} \\
    	\hline 
	\end{tabular}
\end{table*}

\subsection{Results: TCGA Dataset}
\textbf{Dataset Description. } 
We analyze 1-year mortality based on the comprehensive observations from multiple omics on 7,295 cancer cell lines (i.e. samples). The data consists of observations from 4 distinct views on each cell line across 3 different omics layers: (\textbf{View 1}) mRNA expressions, (\textbf{View 2}) DNA methylation, (\textbf{View 3}) microRNA expressions, and (\textbf{View 4}) reverse phase protein array. 
Among 7,295 samples, 3,282 samples have incomplete multi-view observations with various view-missing patterns: average missing rates were 0.10, 0.24, 0.13, and 0.74 for View 1, View 2, View 3, and View 4, respectively. 


\subsubsection{Multi-Omics Data Integration} \label{subsec:tcga_view_integration}
We start off by exploring the benefit of integrating multi-omics data, specifically augmenting samples with incomplete multi-view observations.
To this goal, we compare predictions of different multi-view learning methods in terms of area under the receiver operating characteristics (AUROC) in Table \ref{table:tcga_results} by augmenting $N_{I} = 3,282$ samples with incomplete views on the already available $N_{C} = 3,210$ samples with complete views for training. 
For evaluating the performance, we artificially created missing views for hold-out testing samples by varying the number of observed views from $|\Vc^{n}| = 1$ to $|\Vc^{n}| = 4$.

There are a couple of things to be highlighted from Table \ref{table:tcga_results}: 
First, \proposed~better integrates samples with incomplete views as the performance improvement were most significant outperforming the benchmarks regardless of the number of observed views.
Second, even when trained only with complete-view samples, our method better handles different view missing patterns during testing as it provided the highest performance (except for 1 View) with partially observed views.
Third, MVAE and MOFA sacrifice their discriminative power since the latent representations focus on retaining the information of the input for view generation (reconstruction), which results in discarding the task-relevant discriminative information.

\begin{table}[h!]
	\small
	\centering
	\caption{Comparison of the AUROC performance (mean $\pm$ 95\%-CI) with different view completion methods on the TCGA dataset with $|\Vc^{n}|=3$.}
	\label{table:imputed_results_tcga}
	\begin{tabular}{ l l l }
	\toprule
	\textbf{Methods}&\textbf{Mean Impt.}&\textbf{MVAE Impt.} \\ \midrule
    \textbf{Base1}	  &0.765$\pm$0.02&0.771$\pm$0.01 \\
    \textbf{Base2}	  &0.775$\pm$0.01&0.784$\pm$0.01 \\
    \textbf{GCCA}	  &0.769$\pm$0.02&0.774$\pm$0.01 \\
    \textbf{DCCA}	  &0.775$\pm$0.01&0.784$\pm$0.02 \\
    \textbf{DCCAE} 	  &0.775$\pm$0.01&0.773$\pm$0.02 \\
    \textbf{MVAE}     &0.730$\pm$0.01&$-$\\
    \textbf{CPM-Nets} &0.771$\pm$0.01&$-$ \\
    \textbf{MOFA}     &0.764$\pm$0.02&$-$ \\
    \textbf{DeepIMV}  &\color{blue}\textBF{0.791$\pm$0.01}&$-$ \\
    	\bottomrule
	\end{tabular}
\end{table}
In addition, the proposed method still outperformed the benchmarks that do not handle incomplete views, when we replaced the mean imputation method with an advanced multi-view imputation method (i.e., MVAE) as shown in Table \ref{table:imputed_results_tcga}. (More results can be found in the Supplementary Material.)

\begin{figure*}[t!]
	\centering
	\begin{subfigure}[b]{0.25\linewidth}
		\centering 
		\includegraphics[width=1.0\linewidth, trim= 0.0 10 0.0 0.0, clip]{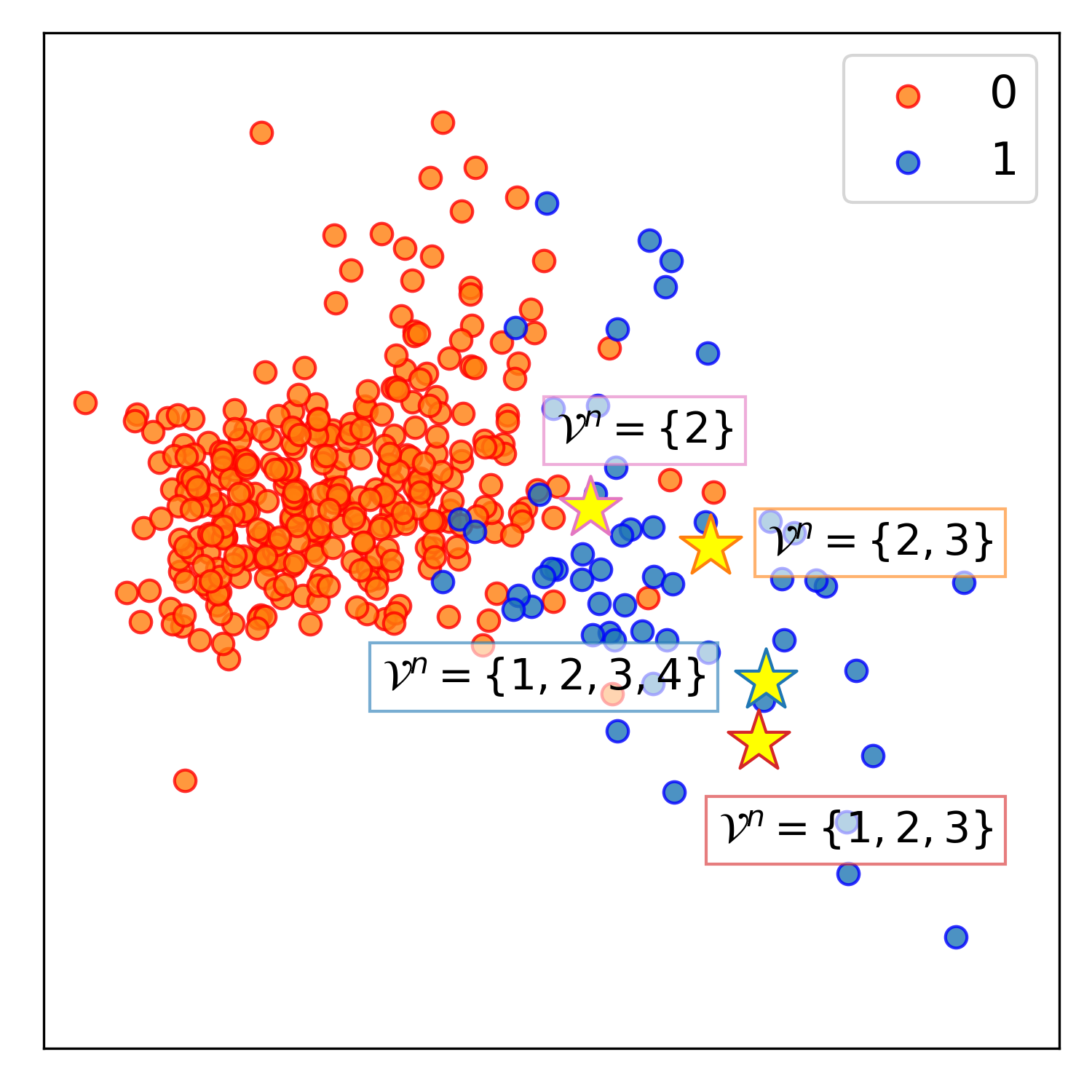}
		\caption{$\Vc = \{1,2,3,4\}$} 
	\end{subfigure}~
	\begin{subfigure}[b]{0.48\linewidth}
		\centering 
		\includegraphics[width=1.0\linewidth, trim= 0.1 0.1 0.1 0.1]{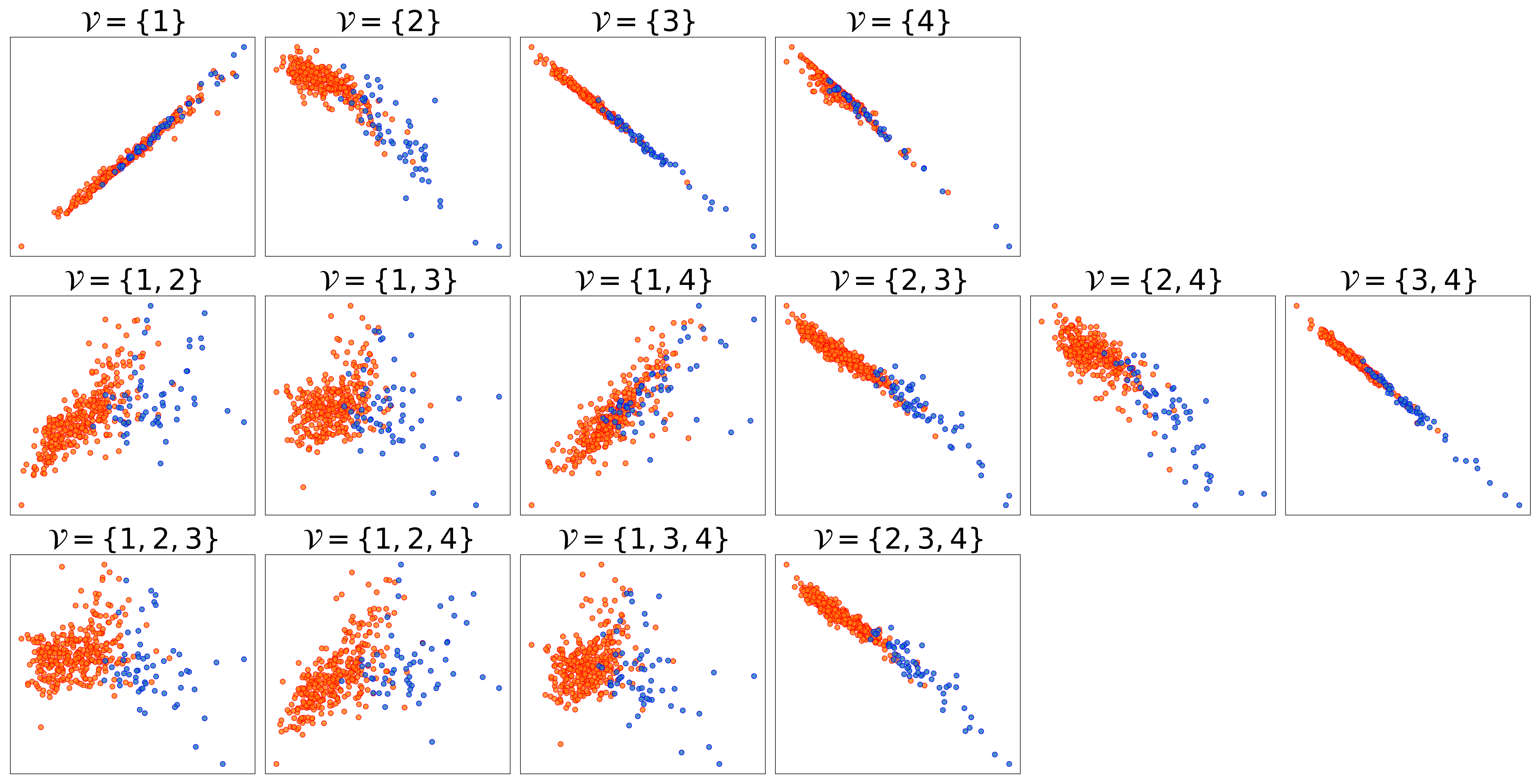}
		\caption{$\Vc \subset \{1,2,3,4\}$}
	\end{subfigure}	
	\caption{Comparison of PCA projections of latent representations from \proposed~trained with both complete and incomplete multi-view samples for the TCGA dataset. Here, all possible combinations of different view-missing patterns $\Vc \subseteq [4]$ are illustrated.}
	\label{fig:pca_comparison}
\end{figure*}

\begin{table*}[t!]
	\small
	\centering
	\caption{Ablation study results. AUROC performances (mean $\pm$ 95\%-CI) are reported for \proposed~variations trained with both complete and incomplete multi-view samples on the TCGA dataset.}
	\label{table:tcga_ablation_study}
	\begin{tabular}{ l  l  l  l  l }
    \toprule
	\textbf{Methods}&\textbf{1 View}&\textbf{2 Views}&\textbf{3 Views}&\textbf{4 Views} \\ \midrule
    MoE &0.629$\pm$0.03&0.691$\pm$0.02&0.736$\pm$0.02&0.768$\pm$0.01 \\
    MoE with marginal IBs&0.712$\pm$0.01&0.766$\pm$0.01&0.786$\pm$0.01&0.790$\pm$0.01 \\
    PoE &0.655$\pm$0.04&0.719$\pm$0.03&0.755$\pm$0.03&0.783$\pm$0.02 \\
    PoE with marginal IBs&\color{blue}\textBF{0.724$\pm$0.02}&\color{blue}\textBF{0.772$\pm$0.01}&\color{blue}\textBF{0.791$\pm$0.01}&\color{blue}\textBF{0.801$\pm$0.01} \\
    \bottomrule
	\end{tabular}
\end{table*}

\textbf{Visualization.}
We visually compare the principle component analysis (PCA) projections of the latent representations of the proposed method with different combinations of observed views in Figure \ref{fig:pca_comparison}. Here, \proposed~is trained with both complete and incomplete multi-view samples.
The PCA projections of latent representations become more discriminative and representative of the target labels as \proposed~incorporates more views during testing. This can be highlighted by how the PCA projections of a representative cell line labeled with 1-year mortality, that are marked by star dots, move toward the opposite direction of the class boundary as the proposed method collects more views from $\Vc^{n} = \{2\}$ (which provides the highest $I(Y_{v};Z_{v})$; see Table \ref{table:ib_quantity_tcga}) to $\Vc^{n} = \{1,2,3,4\}$.
Also, \proposed~was able to achieve very similar representations of complete views without incorporating all the observed views. 
In particular, the PCA projections of latent representations without incorporating View 4 (which provided the smallest $I(Y_{v};Z_{v})$; see Table \ref{table:ib_quantity_tcga}), i.e., $\Vc =\{1,2,3\}$, were almost the same with those of latent representations with $\Vc = \{1,2,3,4\}$. 
This highlights a potential role of \proposed~in designing multi-omics experiments by providing advice which omics layers should not be measured for cost-efficient predictions.
\begin{table}[h!]
	\small
	\centering
	\caption{Information quantities on the TCGA dataset.}
	\label{table:ib_quantity_tcga}
	\begin{tabular}{ c c c c c }
	\toprule
	$I(Y_{1};Z_{1})$ & $I(Y_{2};Z_{2})$ & $I(Y_{3};Z_{3})$ & $I(Y_{4};Z_{4})$ &$I(Y;Z)$ \\
	\midrule
    0.319 & 0.506 & 0.487 & 0.157 & 0.562 \\
	\bottomrule
	\end{tabular}
\end{table}

\subsubsection{Ablation Study} 
In Table \ref{table:tcga_ablation_study}, we study the effect of i) utilizing PoE over MoE, and ii) introducing view-specific predictors and the marginal IB losses in \eqref{eq:loss_IB_marginal} in oder to provide additional insight into the source of gain. 
As discussed in Section \ref{sec:factorization}, a PoE allows for encoders to specialize in analyzing their corresponding views and to build different expertise. 
Furthermore, introducing view-specific predictors and marginal IB losses encourage the view-specific encoders to focus on (possibly complementary) task-relevant information of each view and, thus, to ease the training associated with the PoE factorization.
We observe that these components clearly contributes to \proposed's performance improvement. Here, we use the same experimental setting in Section \ref{subsec:tcga_view_integration}.


\begin{figure*}[t!]
	\centering
	\includegraphics[width=1.00\linewidth, trim= 150 0 150 70, clip]{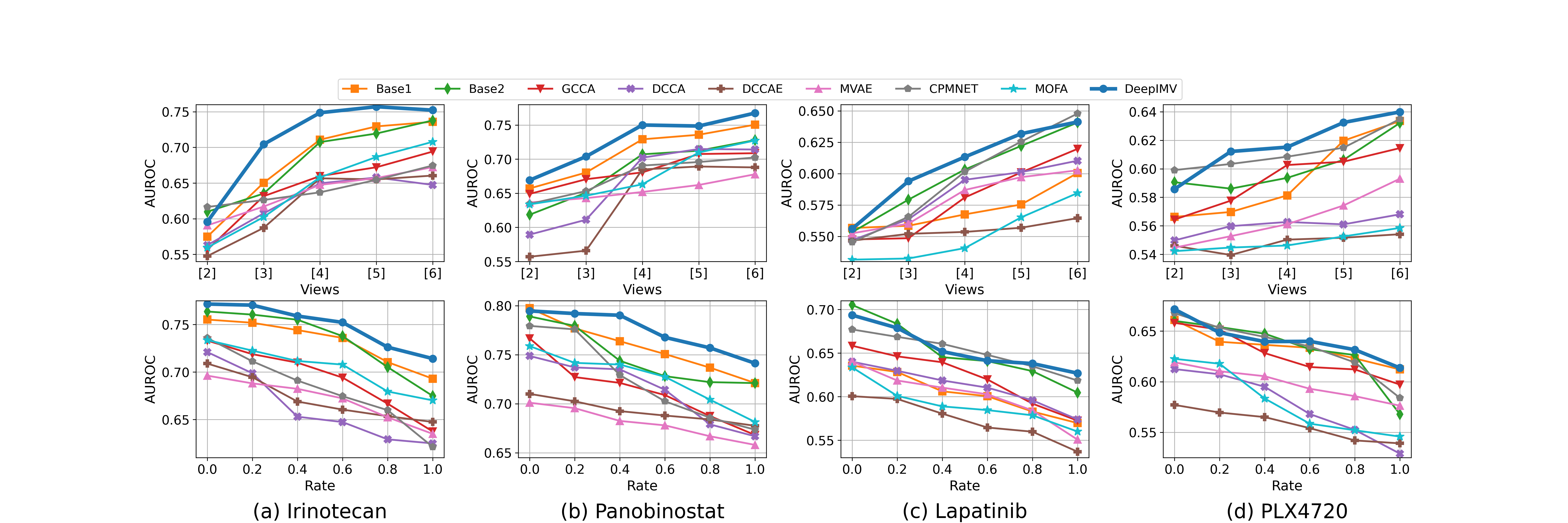}
	\caption{Average AUROC performance on four different drugs in the CCLE dataset (top) with increasing view-missing rates and (bottom) with increasing views.} 
	\label{fig:ccle_all} \vspace{-2mm}
\end{figure*}
\subsection{Results: CCLE Dataset}
\textbf{Dataset Description. } 
We analyze sensitivities of heterogeneous cell lines to 4 different drugs -- that are, \textit{Irinotecan}, \textit{Panobinostat}, \textit{Lapatinib}, and \textit{PLX4720} -- based on the multiple omics observations on 504 cancer cell lines (i.e. samples). Drug response was converted to a binary label by dividing cell lines into quartiles ranked by ActArea; the top 25\% were assigned to the ``sensitive'' class and the rest were assigned to the ``non-sensitive'' class. 
The data consists of observations from 6 distinct views on each cell line across 5 different omics layers: (\textbf{View 1}) DNA copy number, (\textbf{View 2}) DNA methylation, (\textbf{View 3}) mRNA expressions, (\textbf{View 4}) microRNA expressions, (\textbf{View 5}) reverse phase protein array, and (\textbf{View 6}) metabolites. 

\textbf{Incomplete View Construction. }
We artificially construct incomplete multi-view observations based on the following procedures: i) randomly select $N_{I} = N\times R$ samples where $R\in[0,1]$ is the rate of samples with missing views and ii) create incomplete multi-view observations by choosing one of the $(2^{V}-2)$ possible view-missing patterns for each sample. (Obviously, at least one view must be observed).


\subsubsection{Multi-Omics Data Integration} \label{subsec:view_integration}
We explore the benefit of incorporating more samples and more views on predicting drug sensitivities of heterogeneous cell lines. 
To this end, we increase the set of available views from Views $\{1, 2\}$ to Views $\{1, 2, 3, 4, 5, 6\}$ and include samples that have observations from at least one of the views in this set. 
For example, given the set of available views is $\{1,2\}$, we integrate all the samples that satisfy $\Vc^{(n)} \cap \{1,2\} \neq \varnothing$ for training. In the top row of Figure~\ref{fig:ccle_all}, we compare the prediction of different multi-view methods in terms of AUROC performance as we increase the set of available views from $\{1,2\}$ to $\{1, 2, 3, 4, 5, 6\}$ with $R=0.6$.
There are a couple of things to be highlighted from this figure: 
First, our method provides better discriminative performance on all the tested drug sensitivity datasets (most of the times) as the number of integrated views increases. Second, the performances of DCCA and DCCAE are saturated since these methods can utilize only two views at the most, whereas GCCA provides consistently-increasing performance since it generalizes to multiple views. 
Third, MVAE and MOFA sacrifice their discriminative task-relevant information since the latent representations focus on retaining the information of the input for view generation (reconstruction).

\begin{table}[!h]
	\small
	\centering
	\caption{Comparison of the AUROC performance (mean $\pm$ 95\%-CI) with different view completion methods on the CCLE dataset (\textit{Panobinostat}) with $M=6$ and $R=0.6$.}
	\label{table:imputed_results_ccle}
	\begin{tabular}{ l l l }
	\toprule
	\textbf{Methods}&\textbf{Mean Impt.}&\textbf{MVAE Impt.} \\ \midrule
\textbf{Base1}	    &0.751$\pm$0.01&0.758$\pm$0.01 \\
\textbf{Base2}	    &0.728$\pm$0.02&0.752$\pm$0.01 \\
\textbf{GCCA}	    &0.709$\pm$0.01&0.715$\pm$0.01 \\ 
\textbf{DCCA}	    &0.714$\pm$0.01&0.717$\pm$0.01 \\
\textbf{DCCAE}	    &0.688$\pm$0.01&0.697$\pm$0.01 \\
\textbf{MVAE}	    &0.678$\pm$0.01 &$-$ \\
\textbf{CPM-Nets}   &0.702$\pm$0.01 &$-$ \\
\textbf{MOFA}       &0.727$\pm$0.02 &$-$ \\
\textbf{DeepIMV}	&\color{blue}\textBF{0.768$\pm$0.01} &$-$ \\	
    	\bottomrule
	\end{tabular}
\end{table}
In addition, Table \ref{table:imputed_results_ccle} shows that the proposed method still outperformed the benchmarks that do not handle incomplete views, when the mean imputation method is replaced by an advanced multi-view imputation method (i.e., MVAE). (More results can be found in the Supplementary Material.) 

\subsubsection{Robustness to Missing Views}
Next, we evaluate how robust the multi-view learning methods are with respect to the view-missing rate. The bottom row of Figure~\ref{fig:ccle_all} shows the AUROC performance as the rate of samples with missing views ranges from $R=0.0$ (all complete) to $R=1.0$ (all incomplete) with $V=6$. We highlight the following observations: 
First, our method outperforms all the benchmarks on the \textit{Irinotecan} and \textit{Panobinostat} datasets and provides comparable performance on the \textit{Lapatinib} and \textit{PLX4720} datasets to the best performing benchmark across different missing rates.
Second, while other methods often fail, \proposed~provides the most robust performance as the rate of samples with missing views increases. 
Third, DCCA and DCCAE show poor performance since these methods do not fully utilize the available views. 
Last, a similar trend can be found for MVAE and MOFA with that of the previous observation in Section \ref{subsec:view_integration}.

\section{Conclusion}
Our proposed method finds intra-view and inter-view interactions that are relevant for predicting the target labels by flexibly integrating available views regardless of the view missing patterns in a unified framework.
Throughout the experiments, we evaluate \proposed~on real-world multi-omics datasets and show our method significantly outperforms existing multi-view learning methods in terms of prediction performance. 
In the future, further work may investigate incorporating sparsity in different omics data to address high-dimensionality. 


\clearpage

\section*{Acknowledgment}
This work was supported by the National Science Foundation (NSF) (Grant Number: 1722516) and the Office of Naval Research (ONR).

\bibliographystyle{unsrt}

\clearpage
\setcounter{figure}{0}\renewcommand{\thefigure}{S.\arabic{figure}}
\setcounter{table}{0}\renewcommand{\thetable}{S.\arabic{table}}
\setcounter{equation}{0}\renewcommand{\theequation}{S.\arabic{equation}}

	\onecolumn
	\aistatstitle{Supplementary Material for \\
		A Variational Information Bottleneck Approach to Multi-Omics Data Integration}
	
	\appendix
	\section{Details on Variational IB Loss}
	\subsection{Derivation of Variational IB Loss in (1) and (2)}
	Following \cite{sup_Alemi:17}, the variational information bottleneck (IB) loss for the joint latent representation $Z$ can be given as
	\begin{equation} \label{eq:loss_IB_joint} \tag{S.1}
	\loss_{\text{IB-J}}^{\theta,\psi}(\bar{\Xv}, Y) = - I(Z; Y) + \beta I(\bar{\Xv}; Z)
	\end{equation}
	where  $\beta \geq 0$ is a coefficient chosen to balance between the two information quantities. Here, $I(Z; Y)$ and $I(\bar{\Xv}; Z)$ can be derived using variational approximations, i.e., $q_{\theta}(Z|\bar{\Xv})$ and $q_{\psi}(Y|Z)$, as follows:
	\begin{equation} \label{eq:loss_IB_joint_1} \tag{S.2}
	\begin{split}
	I(Z; Y) = \int dzdy~p(z,y) \log \frac{p(y|z)}{p(y)}  &= \int d\bar{\xv}dzdy~p(\bar{\xv},y,z) \log p(y|z) + H(Y) \\
	&= \int d\bar{\xv}dzdy~p(\bar{\xv})p(y|\bar{\xv})p(z|\bar{\xv}) \log p(y|z) + H(Y) \\
	&\approx \int d\bar{\xv}dzdy~p(\bar{\xv})p(y|\bar{\xv})q_{\theta}(z|\bar{\xv}) \log q_{\psi}(y|z) + H(Y) \\
	&= \E_{\bar{\xv},y\sim p(\bar{\xv},y)} \E_{z\sim q_{\theta}(z|\bar{\xv})} \Big[\log q_{\psi}(y|z) \Big] + H(Y)
	\end{split} 
	\end{equation}
	and 
	\begin{equation} \label{eq:loss_IB_joint_2} \tag{S.3}
	\begin{split}
	I(\bar{\Xv}; Z) = \int d\bar{\xv}dz~p(\bar{\xv}, z) \log \frac{p(z|\bar{\xv})}{p(z)} &= \int d\bar{\xv}dz~p(\bar{\xv})p(z|\bar{\xv}) \log \frac{p(z|\bar{\xv})}{p(z)} \\
	&\approx \int d\bar{\xv}dz~p(\bar{\xv})q_{\theta}(z|\bar{\xv}) \log \frac{q_{\theta}(z|\bar{\xv})}{q(z)} \\
	&= \E_{\bar{\xv}\sim p(\bar{\xv})} \Big[ KL\big(q_{\theta}(Z|\bar{\xv}) \big\| q(Z)\big) \Big]
	\end{split} 
	\end{equation}
	where $KL\big(q_{\theta}(Z|\bar{\xv}) \big\| q(Z)\big)$ denotes the Kullback-Leibler (KL) divergence between the two distributions $q_{\theta}(Z|\bar{\xv})$ and $q(Z)$.
	Since the entropy of the labels $H(Y)$ is independent of our optimization procedure, we can simply ignore it and approximate the IB loss by plugging \eqref{eq:loss_IB_joint_1} and \eqref{eq:loss_IB_joint_2} into \eqref{eq:loss_IB_joint}:
	\begin{equation} \label{eq:loss_IB_joint_appx} \nonumber
	\loss_{\text{IB-J}}^{\theta,\psi}(\bar{\Xv}, Y) \approx \E_{\bar{\xv},y\sim p(\bar{\xv},y)} \E_{z\sim q_{\theta}(z|\bar{\xv})} \big[-\log q_{\psi}(y|z) \big] + \beta \E_{\bar{\xv}\sim p(\bar{\xv})} \big[ KL\big(q_{\theta}(Z|\bar{\xv}) \big\| q(Z)\big) \big]
	\end{equation}
	We can similarly derive the IB loss for the marginal representation $Z_{v}$ for $v\in\Vc$.
	
	\subsection{Training \proposed~via IB Losses}
\begin{figure*}[h!]
		\centering
		\begin{subfigure}[b]{1.0\linewidth}
			\centering 
			\includegraphics[width=0.80\linewidth, trim={0cm 0cm 0cm 0cm}, clip]{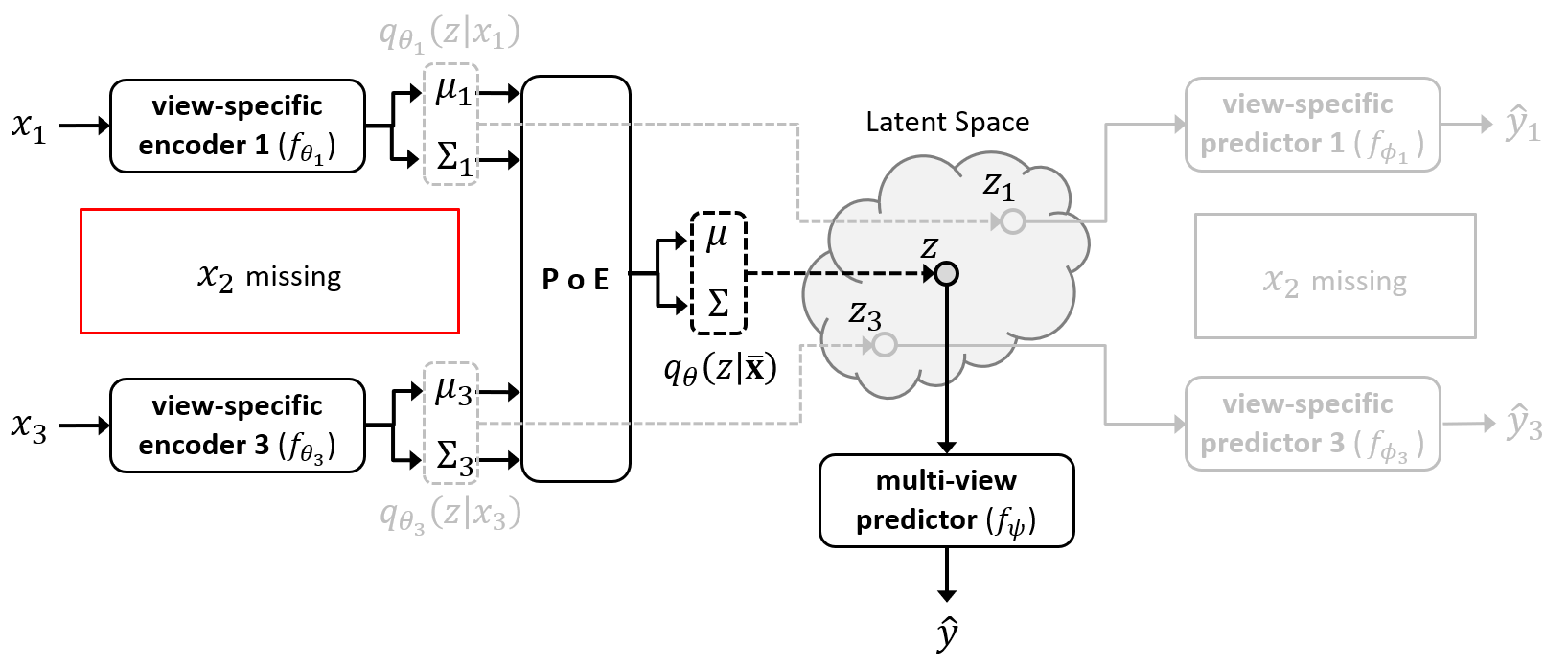} \vspace{-1mm}
			\caption{The network components updated via the joint IB losses in (2)}
		\end{subfigure}
		\begin{subfigure}[b]{1.0\linewidth}
			\centering 
			\includegraphics[width=0.80\linewidth, trim={0cm 0cm 0cm 0cm}, clip]{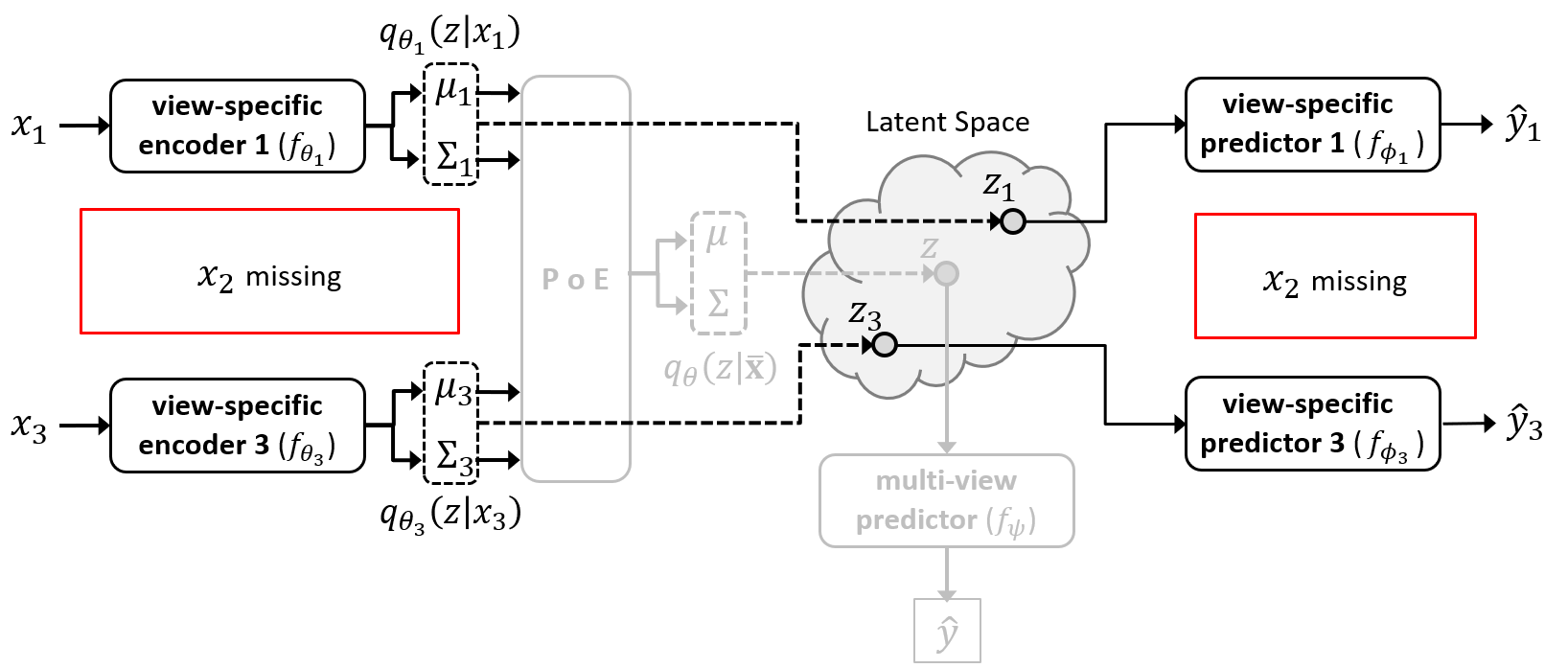} \vspace{-1mm}
			\caption{The network components updated via the marginal IB losses in (1)}
		\end{subfigure}	
		\caption{An illustration of the proposed network architecture with $V=3$ views. For an illustration purpose, we assume that the current sample has the second view missing, i.e., $x_{2}=\varnothing$.}
		\label{fig:trained_network_components}
	\end{figure*}

	Different network components are trained based on the joint and marginal IB losses, respectively. More specifically, the parameters of the view-specific encoders and the multi-view predictor -- i.e., $(\theta, \psi)$ -- are updated based on the joint IB loss while those of the view-specific encoders and the view-specific predictors -- i.e., $(\theta, \phi)$ -- are updated based on the marginal IB losses. Figure \ref{fig:trained_network_components} depicts the network components trained via the joint and the marginal IB losses.

	\section{Implementation Details}
	Among the 4 network components of \proposed, we use multi-layer perceptrons (MLP) as the baseline architecture for the view-specific encoders, view-specific predictors, and multi-view predictor. (Note that the PoE module does not have parameters to be trained.)
	The number of hidden units and layers in each component are optimized by cross-validation, and we choose the ones with the minimum total loss in (4), i.e., $\loss_{\text{IB-J}}^{\theta, \psi}$, on the validation set.  
	The number of hidden units is selected among $\{50, 100, 300\}$ and the number of layers is selected among $\{1, 2, 3\}$. We choose the dimension of latent representations among $\{50, 100\}$ and use $ReLu$ as the activation function at each hidden layer.
	The parameters $(\theta, \phi, \psi)$ are initialized by Xavier initialization \cite{sup_Xavier:10} and optimized via Adam optimizer \cite{sup_Adam:14} with learning rate of $1\times 10^{-4}$. We utilize  dropout \cite{sup_Dropout} with dropout probability $0.7$ to regularize the network.
	The network is further regularized via $L_{1}$-regularization with $1\times 10^{-4}$ for the CCLE dataset and no additional regularization is used for the TCGA dataset.
	
	For the balancing coefficients, we use $\alpha = 1.0$ and $\beta = 0.01$ where we assume that $\beta_{v} = \beta$ for $v \in [V]$ for convenience. Please refer to our sensitivity analysis in Section S.\ref{appx:sensitivity_analysis} for the selection of balancing coefficients $\alpha$ and $\beta$.

	\section{Details of the Benchmarks}
	We compare \proposed~with 2 baseline methods (i.e., Base1 and Base2) and 6 state-of-the-art multi-view learning methods (i.e., GCCA \cite{sup_Kettenring:71}, DCCA \cite{sup_Andrew:13}, DCCAE \cite{sup_Wang:15}, MVAE \cite{sup_Wu:18}, CPM-Nets \cite{sup_Zhang:19}, and MOFA \cite{sup_MOFA}). 
	
	For the baseline methods (i.e., Base1 and Base2), we directly use the observations from multiple views and the corresponding labels for training a MLP. 
	For unsupervised multi-view learning methods (i.e., GCCA, DCCA, DCCAE, MVAE, and MOFA), we use a two-step approach to provide predictions on the target labels: First, we train each method to find the representations of multi-view observations in a common (latent) space. Then, we use the learned representations and the corresponding labels to train a MLP as a downstream task. For the network architecture of MLPs for the downstream task in each benchmark, we use $ReLU$ as the activation function at each hidden layer and use dropout with dropout probability $0.7$ to regularize the network.
	The details of the benchmarks are described as follows: \vspace{-2mm}
	\begin{itemize}[leftmargin=1.5em]
		\item \textbf{Base1}: To handle multi-view observations, we concatenate features from multiple views as a pre-integration step and train a baseline network with the concatenated features as an input. The number of hidden units is selected among $\{50V, 100V, 300V\}$ and the number of layer is selected among $\{1, 2, 3\}$. 
		
		\item \textbf{Base2}: We separately train a MLP for each individual view, and then make an ensemble by averaging the predictions from observed views as a post-integration step. We use a MLP for each view where the number of hidden units is selected among $\{50, 100, 300\}$ and the number of layer is selected among $\{1, 2, 3\}$. 
		
		\item \textbf{GCCA}\footnote{\url{https://github.com/rupy/GCCA}} \cite{sup_Kettenring:71}:  
		GCCA generalizes the CCA framework when there are more than two views. To provide predictions on the target label, we first train GCCA to find the representations in a common space, and then train a MLP based on the concatenated representations. The dimension of the common space is selected among $\{50, 100\}$ which provides the best prediction performance on the validation set. For the downstream task, we use a MLP where the number of hidden units is selected among $\{50, 100, 300\}$ and the number of layer is selected among $\{1, 2, 3\}$. 
		
		\item \textbf{DCCA}\footnote{\url{https://ttic.uchicago.edu/~wwang5/dccae.html} \label{refnote}} \cite{sup_Andrew:13} and \textbf{DCCAE}\footref{refnote} \cite{sup_Wang:15}: DCCA extracts low-dimensional representations for observations from two views in a common space by training neural networks to maximize the canonical correlation between the extracted representations. Similarly, DCCAE extracts low-dimensional representations by training auto-encoders to optimize a combination of reconstruction loss and the canonical correlation. 
		To make predictions on the target task, we first train each method to find latent representations in a latent space, and then train a baseline network based on the concatenated representations.
		For implementation, we utilize MLPs for DCCA and DCCAE. And, we select the number of hidden units among $\{50, 100, 300\}$ and the number of layers among $\{1, 2, 3\}$ based on the validation loss. We use $ReLu$ as the activation function at each hidden layer. The dimension of the latent representation among $\{50, 100\}$ is selected based on the prediction performance on the validation set. 
		
		It is worth highlighting that we select the two best performing views when the available views are more than two, i.e., $V>2$, since DCCA and DCCAE can utilize only two. For the downstream task, we use a MLP where the number of hidden units is selected among $\{50, 100, 300\}$ and the number of layer is selected among $\{1, 2, 3\}$.

		\item \textbf{MVAE}\footnote{\url{https://github.com/mhw32/multimodal-vae-public}} \cite{sup_Wu:18}: MVAE learns latent representations for incomplete multi-view observations that can generate the original views under the VAE framework. We modify the publicly available code since it only supports observations from two views. We implement the VAE components using MLPs where the number of hidden units, the number of layers, and the dimension of the latent representations are selected among $\{50, 100, 300\}$, $\{1, 2, 3\}$, and $\{50, 100\}$, respectively, based on its validation loss. We use $ReLu$ as the activation function at each hidden layer. To make predictions on the target task, we first train MVAE to find latent representations of incomplete multi-view observations in the latent space, and then train a MLP based on the learned representations. For the downstream task, the number of hidden units is selected among $\{50, 100, 300\}$ and the number of layer is selected among $\{1, 2, 3\}$. 
		
		\item \textbf{CPM-Nets}\footnote{\url{{https://github.com/hanmenghan/CPM_Nets}}} \cite{sup_Zhang:19}: CPM-Nets learns representations in a common space to provide predictions on the target classification task based on the incomplete multi-view observations. We implement each component of CPM-Nets using MLPs where the number of hidden units, the number of layers, and the dimension of the latent representations are selected among $\{50, 100, 300\}$, $\{1, 2, 3\}$, and $\{50, 100\}$, respectively, based on the prediction performance on the validation set. We use $ReLu$ as the activation function at each hidden layer. 
		
		\item \textbf{MOFA}\footnote{\url{https://pypi.org/project/mofapy/}} \cite{sup_MOFA}: MOFA infers a low-dimensional representation of the data in terms of a small number of (latent) factors that capture the joint aspects across different views. For training MOFA, we used the original views (i.e., views without conducting kernel PCA) as it is well-known for capturing sparse factors across multiple views. We set the initial number of factors as $50$. For the downstream task, the number of hidden units is selected among $\{50, 100, 300\}$ and the number of layer is selected among $\{1, 2, 3\}$. 
	\end{itemize} \vspace{-2mm}
	
	It is worth highlighting that among the benchmarks, MVAE, CPM-Nets, and MOFA can flexibly handle incomplete multi-view observations during training. Hence, for training the other benchmarks (except for Base2), we use mean imputation for missing views. 
	For Base2, we train the baseline network for each individual view using samples that have observations for the corresponding view.

	\section{Obtaining Multi-Omics Datasets}
	\subsection{TCGA Dataset}
	For constructing multiple views and the labels, the following datasets were downloaded from 
	\url{http://gdac.broadinstitute.org}:
	\begin{itemize}[leftmargin=1.5em]
		\item DNA methylation (epigenomics): \textit{Methylation\_Preprocess.Level\_3.2016012800.0.0.tar.gz}
		\item microRNA expression (transcriptomics): \textit{miRseq\_Preprocess.Level\_3.2016012800.0.0.tar.gz}
		\item mRNA expression (transcriptomics): \textit{mRNAseq\_Preprocess.Level\_3.2016012800.0.0.tar.gz}
		\item RPPA (proteomics): \textit{RPPA\_AnnotateWithGene.Level\_3.2016012800.0.0.tar.gz}
		\item clinical labels: \textit{Clinical\_Pick\_Tier1.Level\_4.2016012800.0.0.tar.gz}
	\end{itemize}
	Time to death or censoring in clinical labels was converted to a binary label for 1-year mortality.
	
	\subsection{CCLE Dataset}
	For constructing multiple views and the labels, the following datasets were downloaded from \url{https://portals.broadinstitute.org/ccle/data}:
	\begin{itemize}[leftmargin=1.5em]
		\item DNA copy number (genomics): \textit{CCLE\_copynumber\_byGene\_2013-12-03.txt}
		\item DNA methylation (epigenomics): \textit{CCLE\_RRBS\_enh\_CpG\_clusters\_20181119.txt}
		\item microRNA expression (transcriptomics): \textit{CCLE\_miRNA\_20181103.gct.txt}
		\item mRNA expression (transcriptomics): \textit{CCLE\_RNAseq\_genes\_counts\_20180929.gct}
		\item RPPA (proteomics): \textit{CCLE\_RPPA\_20181003.csv}
		\item metabolites (Metabolomics):\textit{CCLE\_metabolomics\_20190502.csv}
		\item drug sensitivities: \textit{CCLE\_NP24.2009\_Drug\_data\_2015.02.24.csv}
	\end{itemize}
	Drug response was converted to a binary label by dividing cell lines into quartiles ranked by ActArea; the top 25\% were assigned to the ``sensitive'' class and the rest were assigned to the``non-sensitiv'' class.
	
	For both datasets, we imputed missing values within the observed views with mean values. To focus our experiments on the integrative analysis and to avoid ``curse-of-dimensionality'' in the high-dimensional multi-omics data, we extracted low-dimensional representations (i.e., 100 features) using the kernel-PCA (with polynomial kernels) on each view \cite{sup_Shiokawa:18}.

	\section{Additional Experiments}
	
	\begingroup
	\setlength{\tabcolsep}{5.5pt} 
	\begin{table*}[t!]
		\scriptsize
		\centering
		\caption{Comparison of the AUROC performance (mean $\pm$ 95\%-CI) with different view completion methods on the TCGA dataset. All methods are trained with both complete and incomplete multi-view samples. The values are reported by varying the number of observed views from $|\Vc^{n}| = 1$ to $|\Vc^{n}| = 4$ during testing.}
		\label{table:imputed_results_tcga_all}
		\begin{tabular}{| c | c c | c c | c c | c c |}
			\hline 
			\multirow{2}{*}{\textbf{Methods}} &\multicolumn{2}{c|}{\textbf{\underline{~~~~~\textit{1 View}~~~~~}}} & \multicolumn{2}{c|}{\textbf{\underline{~~~~~\textit{2 Views}~~~~~}}} & \multicolumn{2}{c|}{\textbf{\underline{~~~~~\textit{3 Views}~~~~~}}}  &
			\multicolumn{2}{c|}{\textbf{\underline{~~~~~\textit{4 Views}~~~~~}}} \\ 
			&\textit{mean impt.} &\textit{MVAE impt.} &\textit{mean impt.} &\textit{MVAE impt.} &\textit{mean impt.} &\textit{MVAE impt.} &\textit{mean impt.} &\textit{MVAE impt.}\\ \hline 
			\textbf{Base1}	&0.675$\pm$0.02&0.679$\pm$0.02&0.739$\pm$0.02&0.744$\pm$0.01&0.765$\pm$0.02&0.771$\pm$0.01&0.781$\pm$0.01&0.780$\pm$0.02 \\
			\textbf{Base2}	&0.717$\pm$0.02&0.717$\pm$0.02&0.766$\pm$0.00&0.765$\pm$0.02&0.775$\pm$0.01&0.784$\pm$0.01&0.790$\pm$0.01&0.790$\pm$0.01 \\
			\textbf{GCCA}	&0.650$\pm$0.03&0.660$\pm$0.01&0.737$\pm$0.03&0.737$\pm$0.03&0.769$\pm$0.02&0.774$\pm$0.01&0.792$\pm$0.01&0.794$\pm$0.00 \\
			\textbf{DCCA}	&0.638$\pm$0.03&0.671$\pm$0.02&0.761$\pm$0.02&0.763$\pm$0.02&0.775$\pm$0.01&0.784$\pm$0.02&0.784$\pm$0.01&0.794$\pm$0.01 \\
			\textbf{DCCAE}	&0.605$\pm$0.04&0.626$\pm$0.03&0.763$\pm$0.01&0.763$\pm$0.03&0.775$\pm$0.01&0.773$\pm$0.02&0.778$\pm$0.02&0.779$\pm$0.01 \\
			\textbf{MVAE}    &\multicolumn{2}{c|}{0.589$\pm$0.04}&\multicolumn{2}{c|}{0.674$\pm$0.02}&\multicolumn{2}{c|}{0.730$\pm$0.01}&\multicolumn{2}{c|}{0.781$\pm$0.01} \\
			\textbf{CPM-Nets}&\multicolumn{2}{c|}{0.709$\pm$0.01}&\multicolumn{2}{c|}{0.761$\pm$0.02}&\multicolumn{2}{c|}{0.771$\pm$0.01}&\multicolumn{2}{c|}{0.788$\pm$0.01} \\
			\textbf{DeepIMV} &\multicolumn{2}{c|}{\color{blue}\textBF{0.724$\pm$0.02}}&\multicolumn{2}{c|}{\color{blue}\textBF{0.772$\pm$0.01}}&\multicolumn{2}{c|}{\color{blue}\textBF{0.791$\pm$0.01}}&\multicolumn{2}{c|}{\color{blue}\textBF{0.801$\pm$0.01}}\\
			\hline 
		\end{tabular}
	\end{table*}
	\endgroup
	\begingroup
	\setlength{\tabcolsep}{5.5pt} 
	\begin{table*}[t!]
		\scriptsize
		\centering
		\caption{
			Comparison of the AUROC performance (mean $\pm$ 95\%-CI) with different view completion methods on the CCLE dataset with $M=6$ and $R=0.6$.}
		\label{table:imputed_results_ccle_all}
		\begin{tabular}{| c | c c | c c | c c | c c |}
			\hline 
			\multirow{2}{*}{\textbf{Methods}} &\multicolumn{2}{c|}{\textbf{\underline{~~~~~\textit{Irinotecan}~~~~~}}} & \multicolumn{2}{c|}{\textbf{\underline{~~~~~\textit{Panobinostat}~~~~~}}} & \multicolumn{2}{c|}{\textbf{\underline{~~~~~\textit{Lapatinib}~~~~~}}}  &
			\multicolumn{2}{c|}{\textbf{\underline{~~~~~\textit{PLX4720}~~~~~}}} \\ 
			&\textit{mean impt.} &\textit{MVAE impt.} &\textit{mean impt.} &\textit{MVAE impt.} &\textit{mean impt.} &\textit{MVAE impt.} &\textit{mean impt.} &\textit{MVAE impt.}\\ \hline 
			\textbf{Base1}	&0.736$\pm$0.01&0.726$\pm$0.02&0.751$\pm$0.01&0.758$\pm$0.01&0.600$\pm$0.01&0.632$\pm$0.01&0.633$\pm$0.01&0.630$\pm$0.01 \\
			\textbf{Base2}	&0.738$\pm$0.02&0.730$\pm$0.02&0.728$\pm$0.02&0.752$\pm$0.01&0.641$\pm$0.01&0.627$\pm$0.01&0.632$\pm$0.02&0.631$\pm$0.02 \\
			\textbf{GCCA}	&0.694$\pm$0.01&0.698$\pm$0.02&0.709$\pm$0.01&0.715$\pm$0.01&0.620$\pm$0.01&0.619$\pm$0.01&0.615$\pm$0.02&0.617$\pm$0.01 \\
			\textbf{DCCA}	&0.647$\pm$0.02&0.662$\pm$0.02&0.714$\pm$0.01&0.717$\pm$0.01&0.610$\pm$0.01&0.608$\pm$0.02&0.568$\pm$0.02&0.567$\pm$0.01 \\
			\textbf{DCCAE}	&0.661$\pm$0.02&0.660$\pm$0.02&0.688$\pm$0.01&0.697$\pm$0.01&0.565$\pm$0.01&0.559$\pm$0.01&0.554$\pm$0.01&0.563$\pm$0.01 \\
			\textbf{MVAE}	&\multicolumn{2}{c|}{0.672$\pm$0.01}&\multicolumn{2}{c|}{0.678$\pm$0.01}&\multicolumn{2}{c|}{0.603$\pm$0.01}&\multicolumn{2}{c|}{0.593$\pm$0.01} \\
			\textbf{CPM-Nets}&\multicolumn{2}{c|}{0.675$\pm$0.02}&\multicolumn{2}{c|}{0.702$\pm$0.01}&\multicolumn{2}{c|}{\color{blue}\textBF{0.648$\pm$0.01}}&\multicolumn{2}{c|}{0.635$\pm$0.01} \\	
			\textbf{MOFA}&\multicolumn{2}{c|}{0.708$\pm$0.02}&\multicolumn{2}{c|}{0.727$\pm$0.02}&\multicolumn{2}{c|}{0.585$\pm$0.02}&\multicolumn{2}{c|}{0.559$\pm$0.02} \\
			\textbf{DeepIMV}	&\multicolumn{2}{c|}{\color{blue}\textBF{0.752$\pm$0.01}}&\multicolumn{2}{c|}{\color{blue}\textBF{0.768$\pm$0.01}}&\multicolumn{2}{c|}{0.641$\pm$0.01}&\multicolumn{2}{c|}{\color{blue}\textBF{0.640$\pm$0.01}} \\	
			\hline 
		\end{tabular}
	\end{table*}
	\endgroup
	\subsection{Additional Experiments with Multi-View Imputations}
	We also imputed observations from missing views by utilizing the reconstructed inputs of MVAE, which can flexibly integrate incomplete multi-view observations regardless of the view-missing patterns. 
	Table \ref{table:imputed_results_tcga_all} and \ref{table:imputed_results_ccle_all} shows the AUROC performance when two different imputation methods are used for the multi-view learning methods (except for MVAE, CPM-Nets, MOFA, and \proposed~that do not depend on the imputation methods), for the TCGA dataset and the CCLE dataset, respectively. For the TCGA dataset, all the methods are trained with both complete-view and incomplete-view samples. And, for the CCLE dataset, we set $M=6$ and $R=0.6$ for constructing missing views. 
	The benchmarks trained with imputed observations based on MVAE did not always provide performance gain over those trained with mean imputed observations since reconstructing the inputs can fail to maintain information that is relevant for predicting the target. 
	Even when the imputation based on MVAE improves the discriminative performance, our method still outperforms the benchmark for all the datasets except for \textit{Lapatinib} of the CCLE dataset.

	\subsection{Sensitivity Analysis -- Effects of $\alpha$ and $\beta$} \label{appx:sensitivity_analysis}
\begin{figure*}[h!]
		\centering
		\begin{subfigure}[b]{0.49\linewidth}
			\centering 
			\includegraphics[width=1.00\linewidth, trim= 0.1 0.1 0.1 0.1]{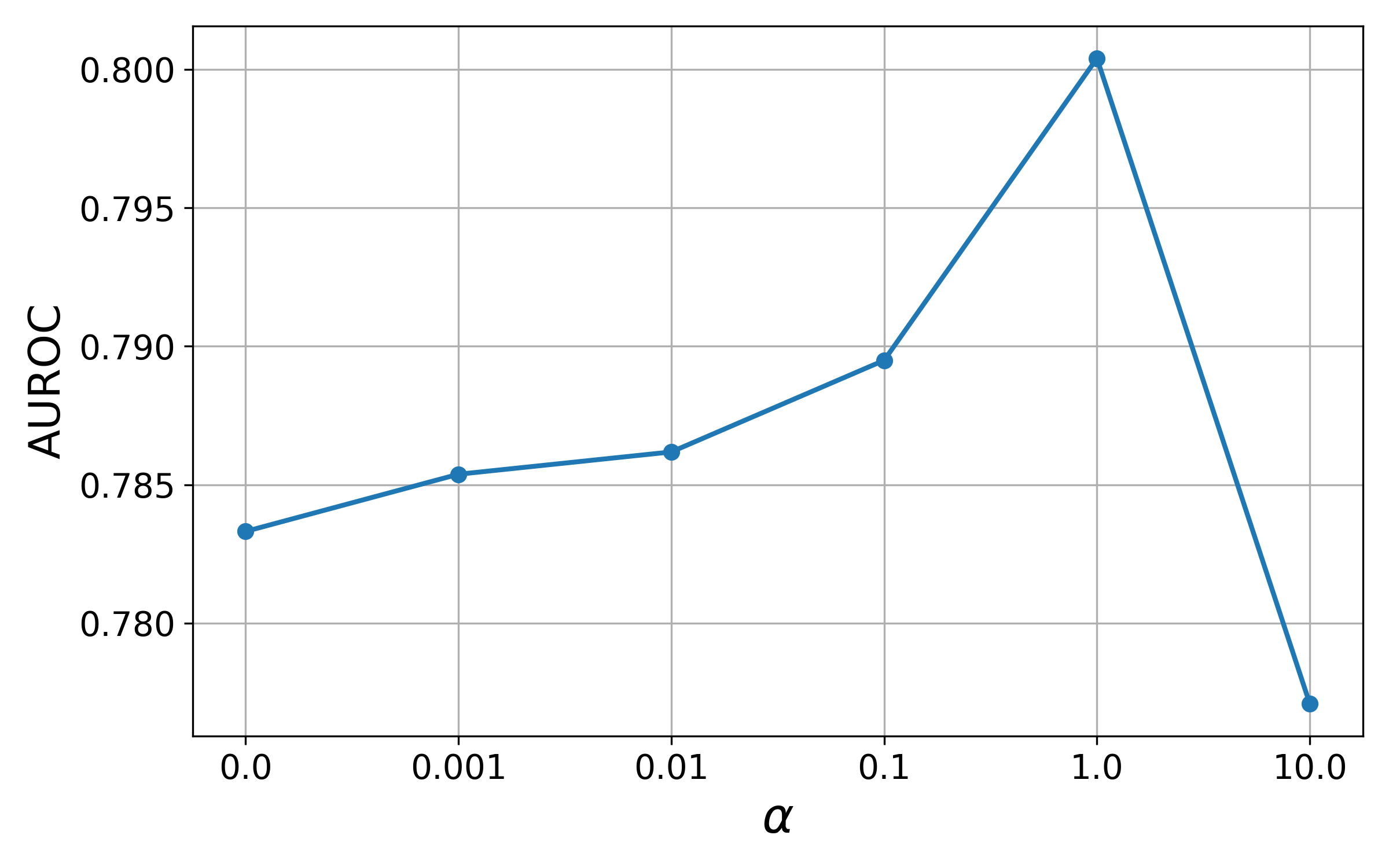}
			\caption{AUROC vs $\alpha$} \label{fig:sensitivity_analysis_alpha}
		\end{subfigure}	
		\begin{subfigure}[b]{0.49\linewidth}
			\centering 
			\includegraphics[width=1.00\linewidth, trim= 0.1 0.1 0.1 0.1]{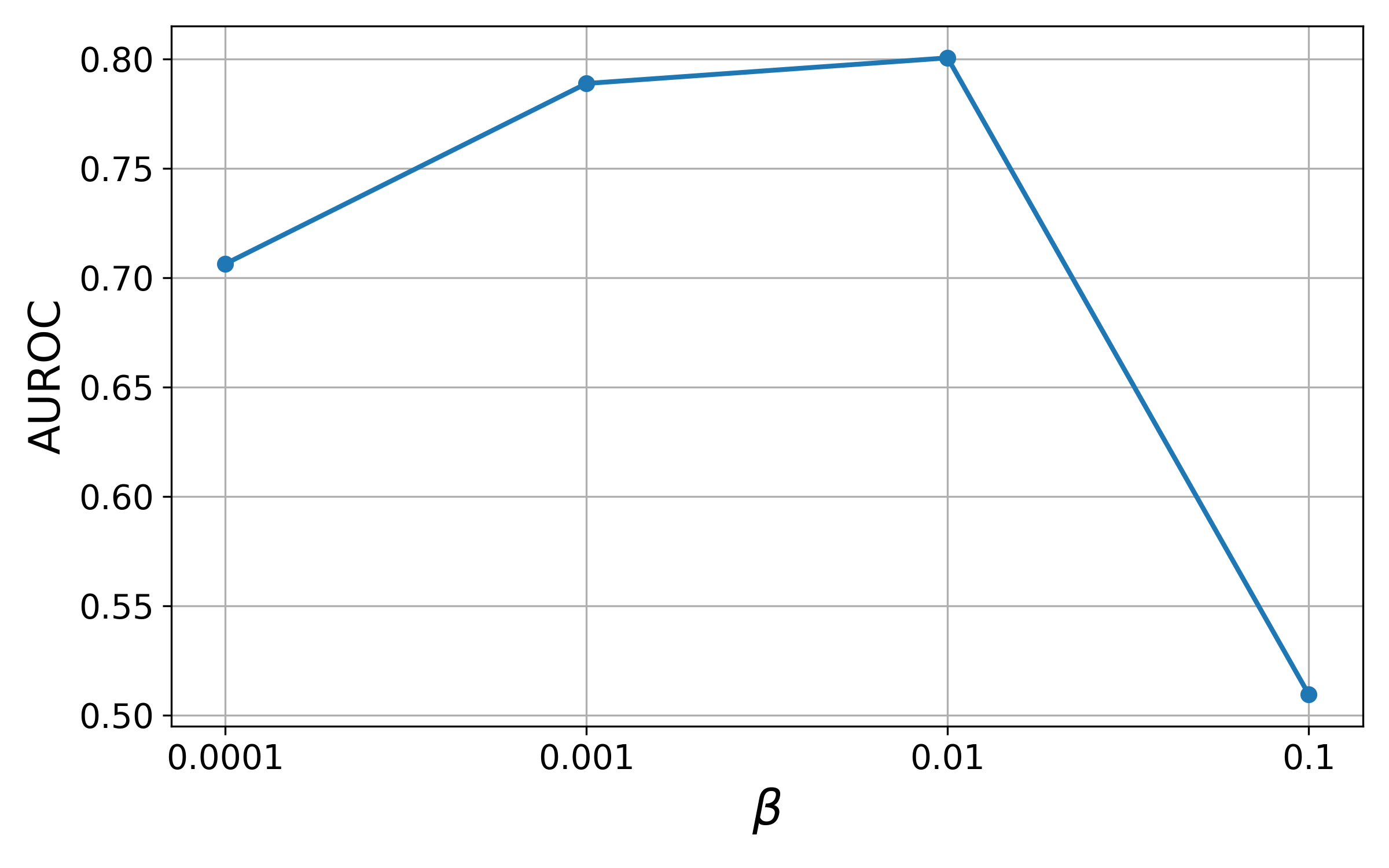}
			\caption{AUROC vs $\beta$} \label{fig:sensitivity_analysis_beta}
		\end{subfigure}	
		\caption{The AUROC performance for the TCGA dataset with $|\Vc^{n}| = 4$ in terms of different values for (a) $\alpha$ (while setting  $\beta = 0.01$) and (b) $\beta$ (while setting  $\alpha = 1.0$).} 
		\label{fig:sensitivity_analysis}
	\end{figure*}

	In this section, we provide sensitivity analysis using the TCGA dataset to see effects of $\alpha$ and $\beta$ on the prediction performance of \proposed.
	Figure \ref{fig:sensitivity_analysis} shows the AUROC performance of our method with respect to different values of $\alpha$ and $\beta$, respectively. For training the variants of the proposed method, we used both complete-view and incomplete-view samples.
	
	\textbf{The Effect of $\alpha$. }
	As shown in Figure \ref{fig:sensitivity_analysis_alpha}, the discriminative performance drops at $\alpha = 10.0$ since too high value of $\alpha$ makes \proposed~to focus on view-specific aspects which may end up with sacrificing joint aspects of observed views for predicting the target. Contrarily, too small value of $\alpha$ makes the learning of task-relevant information from the observed views difficult since each marginal representations do not capture the important information for predicting the target from the corresponding view.

	\textbf{The Effect of $\beta$. }
	Similar to the findings via extensive experiments in \cite{sup_Alemi:17}, $\beta$, which balances between having a representation that is concise and one that provides good prediction power, plays an important role in \proposed.  
	As shown in Figure \ref{fig:sensitivity_analysis_beta}, the classification performance drops at $\beta = 0.1$ since too high value of $\beta$ blocks information from the input that is required to provide good predictions on the target task. 
	For small values of $\beta$, we witness \proposed~becomes overfitted since the view-specific encoder block learns to be more deterministic and thereby reducing the benefits of regularization. (Note that, in such cases, early-stopping is used to prevent from overfitting.)
	
	Throughout our experiments, we set $(\alpha,\beta) = (1.0, 0.01)$ for the TCGA dataset and $(\alpha, \beta) = (0.1, 0.01)$ for the CCLE dataset.

	\clearpage
	\bibliographystyle{unsrt}

	\vfill

\end{document}